\documentclass{article}
\usepackage[]{arxiv}

\usepackage[utf8]{inputenc} 
\usepackage[T1]{fontenc}    
\usepackage{hyperref}       
\usepackage{url}            
\usepackage{booktabs}       
\usepackage{amsfonts}       
\usepackage{nicefrac}       
\usepackage{microtype}      

\usepackage{multirow}
\usepackage{multicol}
\usepackage{caption}
\usepackage{subcaption}
\usepackage{graphicx}
\usepackage{dsfont}
\usepackage{amsmath}
\usepackage{amsthm}
\usepackage{stackengine}
\usepackage{amssymb}
\usepackage{algorithm}
\usepackage{algorithmicx}
\usepackage{algpseudocode} 
\usepackage{makecell}
\usepackage{color}
\usepackage{todonotes}
\usepackage{caption}
\usepackage{subcaption}
\usepackage{pbox}

\usepackage{tikz}
\usepackage{pgfplots, pgfplotstable}
\pgfplotsset{compat=1.9}

\usepackage{mathabx}

\newcommand{\True}{\mbox{1}}

\algnewcommand{\LineComment}[1]{\State \(\triangleright\) #1}
\usepackage{mathtools}
\DeclarePairedDelimiter\ceil{\lceil}{\rceil}

\hypersetup{
    colorlinks=true,
    linkcolor=blue,
    filecolor=blue,      
    urlcolor=blue,
    citecolor=.
}

\title{The Convolutional Tsetlin Machine\thanks{Source code and demos for this paper can be found at \href{https://github.com/cair/pyTsetlinMachineParallel}{https://github.com/cair/pyTsetlinMachineParallel}.}}

\author{
 Ole-Christoffer Granmo\\
 CAIR, University of Agder\\
  \And
  Sondre Glimsdal\\
  CAIR, University of Agder\\
  \And
  Lei Jiao\\
  CAIR, University of Agder\\
  \And
  Morten Goodwin\\
  CAIR, University of Agder\\
  \And
  Christian W. Omlin\\
  CAIR, University of Agder\\
  \And
  Geir Thore Berge\\
  SSHF and CAIR\\
}

\begin{document}

\maketitle

\begin{abstract}
Convolutional neural networks (CNNs) have obtained astounding successes for important pattern recognition tasks, but they suffer from high computational complexity and the lack of interpretability. The recent Tsetlin Machine (TM) attempts to address this lack by using easy-to-interpret \emph{conjunctive clauses} in propositional logic to solve complex pattern recognition problems. The TM provides competitive accuracy in several benchmarks, while keeping the important property of interpretability. It further facilitates hardware-near implementation since inputs, patterns, and outputs are expressed as bits, while recognition and learning rely on straightforward bit manipulation. In this paper, we exploit the TM paradigm by introducing the \emph{Convolutional Tsetlin Machine} (CTM), as an interpretable alternative to CNNs. Whereas the TM categorizes an image by employing each clause once to the whole image, the CTM uses each clause as a convolution filter. That is, a clause is evaluated multiple times, once per image patch taking part in the convolution. To make the clauses location-aware, each patch is further augmented with its coordinates within the image. The output of a convolution clause is obtained simply by ORing the outcome of evaluating the clause on each patch. In the learning phase of the TM, clauses that evaluate to $1$ are contrasted against the input. For the CTM, we instead contrast against one of the patches, \emph{randomly} selected among the patches that made the clause evaluate to $1$. Accordingly, the standard Type I and Type II feedback of the classic TM can be employed directly, without further modification. The CTM obtains a peak test accuracy of $99.4$\% on MNIST, $96.31$\% on Kuzushiji-MNIST, $91.5$\% on Fashion-MNIST, and $100.0$\% on the 2D Noisy XOR Problem, which is competitive with results reported for simple 4-layer CNNs, BinaryConnect, Logistic Circuits, and a recent FPGA-accelerated Binary CNN. 
\end{abstract}

\section{Introduction}
The Tsetlin Machine (TM) \cite{granmo2018tsetlin} is a novel  machine learning paradigm introduced in 2018. It is based on the Tsetlin Automaton (TA) \cite{Tsetlin1961}, one of the pioneering solutions to the well-known multi-armed bandit problem \cite{Robbins1952,Gittins1979} and the first Finite State Learning Automaton (FSLA) \cite{Narendra1989}. The TM has the following main properties that make it attractive as a building block for machine learning~\cite{granmo2018tsetlin}: (a) it solves complex pattern recognition problems with interpretable propositional formulae. This is crucial for high stakes decisions \cite{Rudin2019}. (b) It learns on-line, persistently increasing both training- and test accuracy, before converging to a Nash equilibrium. The Nash equilibrium balances false positive against false negative classifications, while combating overfitting with frequent pattern mining principles. (c) Resource allocation dynamics are leveraged to optimize usage of limited pattern representation resources. By allocating resources uniformly across sub-patterns, local optima are avoided. (d) Inputs, patterns, and outputs are expressed as bits, while recognition and learning rely on straightforward bit manipulation. This facilitates low-energy consuming hardware design \cite{wheeldon2020hardware}. (e) Finally, the TM has provided competitive accuracy in comparison with classical and neural network based techniques, while keeping the important property of interpretability \cite{berge2019,abeyrathna2019nonlinear}.

The TM currently is state-of-the-art in FSLA-based pattern recognition. However, when compared with CNNs, it struggles with attaining competitive accuracy, providing e.g. $98.5$\% mean accuracy on MNIST (without augmenting the data) \cite{granmo2018tsetlin}. To address this deficiency, we here introduce the Convolutional Tsetlin Machine (CTM), a new kind of TM designed for image classification. 

\textbf{FSLA.} The simple Tsetlin Automaton approach has formed the core of more advanced FSLA designs that solve a wide range of problems. This includes resource allocation \cite{Granmo2010g},
decentralized control \cite{Tung1996},
knapsack problems \cite{Granmo2007d}, searching on the line \cite{Oommen1997}, meta-learning \cite{Oommen2008}, the satisfiability problem \cite{Granmo2007c}, graph colouring \cite{Bouhmala2010}, preference learning \cite{Yazidi2012b}, frequent itemset mining \cite{Haugland2014}, adaptive sampling \cite{Granmo2010},
spatio-temporal event detection \cite{Yazidi2013},
equi-partitioning \cite{Oommen1988}, streaming sampling for social activity networks \cite{Ghavipour2018}, routing bandwidth-guaranteed paths \cite{Oommen2007a},
faulty dichotomous search \cite{Yazidi2018}, and learning in deceptive environments \cite{Zhang2016a}, to list a few examples. The unique strength of all of these FSLA designs is that they provide state-of-the-art performance when problem properties are unknown and stochastic, and the problem must be solved as quickly as possible through trial and error. 

\textbf{Rule-based Machine Learning.}
While the present paper focuses on extending the field of FSLA, we acknowledge the extensive work on rule-based interpretable pattern recognition from other fields of machine learning. Learning propositional formulae to represent patterns in data has a long history. One prominent example is frequent itemset mining for association rule learning~\cite{Agrawal1993}, for instance applied to predicting sequential events \cite{rudin13,mccormick15}.  Other examples include the work of Feldman who investigated the hardness of learning formulae in disjunctive normal form (DNF) \cite{feldman9}. Furthermore, Probably Approximately Correct (PAC) learning has provided fundamental insight into machine learning, as well as a framework for learning formulae in DNF \cite{valiant12}. Approximate Bayesian approaches have recently been introduced to provide more robust learning of rules \cite{wang6,hauser13}.  However, in general, rule-based machine learning scales poorly and is prone to noise. Indeed, for data-rich problems, in particular those involving natural language and sensory inputs, state-of-the-art rule-based machine learning is inferior to deep learning.
The recent hybrid Logistic Circuits \cite{LiangAAAI19}, however, have had success in image classification. This approach uses local search to build Bayesian models that capture logical expressions, and learns to classify by employing stochastic gradient descent.
The CTM, on the other hand, attempts to bridge the gap between the interpretability of pure rule-based machine learning and the accuracy of deep learning, by allowing the TM to more effectively deal with images.

\textbf{CNNs.} A myriad of image recognition techniques have been reported. However, after AlexNet won the ImageNet recognition challenge by a significant margin in 2012, the entire field of computer vision has been dominated by CNNs. The AlexNet architecture was built upon earlier work by LeCun et al. \cite{lecun1998gradient}  who introduced CNNs in 1998. Countless CNN architectures, all following the same basic principles, have since been published, including the now state-of-the-art Squeeze-and-Excitation networks  \cite{hu2018squeeze}. In this paper, we introduce convolution to the TM paradigm of machine learning. By doing so, we simultaneously propose a new kind of convolution filter: an interpretable filter expressed as a  propositional formula. Our intent is to address a well-known disadvantage of CNNs, namely, that the CNN models in general are complex and non-transparent, making them hard to interpret. Consequently, the knowledge on why CNNs perform so well and what steps are need to improve the models is limited \cite{zeiler2014visualizing}.

\textbf{Binary CNNs.} Relying on a large number of multiply-accumulate operations, training CNNs is computationally intensive. To mitigate this, there is an increasing interest in binarizing the CNNs. With only two possible values for the synapse weights, e.g., $-1$ and $1$, many of the multiplication operations can be replaced with simple accumulations. This could potentially open up for more specialized hardware and more compressed and efficient models. One recent approach is BinaryConnect \cite{courbariaux2015binaryconnect}, which reached a near state-of-the-art accuracy of $98.99$\% on MNIST, and $99.13$\% in combination with an SVM. Binary CNNs have been further improved with the introduction of XNOR-Nets, which replace the standard CNN filters with binary equivalents \cite{rastegari2016xnor}. BNN+ \cite{Darabi2018} is the most recent binary CNN, extending the XNOR-Nets with two additional regularization functions and an adaptive scaling factor. The CTM can be seen as an extreme Binary CNN in the sense that it is entirely based on fast summation and logical operators. Additionally, learning in the CTM is bandit based (online learning from reinforcement), while Binary CNNs are based on backpropagation.

\textbf{Contributions and Paper Outline.} Our contributions can be summarized as follows. First, in Sect. \ref{sec:TM}, we provide a brief introduction to the TM and a succinct definition of both recognition and learning. Then, in Sect. \ref{sec:CTM}, we introduce the concept of convolution to TMs. Whereas the classic TM categorizes an image by employing each clause once on the whole image, the CTM uses each clause as a convolution filter. In all brevity, we propose a recognition and learning approach for clause-based convolution that produces interpretable filters. In Sect. \ref{sec:empirical_results}, we evaluate the CTM on MNIST, Kuzushiji-MNIST, Fashion-MNIST, and the 2D Noisy XOR Problem, and discuss the merits of the new scheme. Finally, we conclude and provide pointers to further work in Sect. \ref{sec:conclusion}.

\section{The Tsetlin Machine}
\label{sec:TM}

\begin{figure}[ht]
\centering
\includegraphics[width=0.9\textwidth]{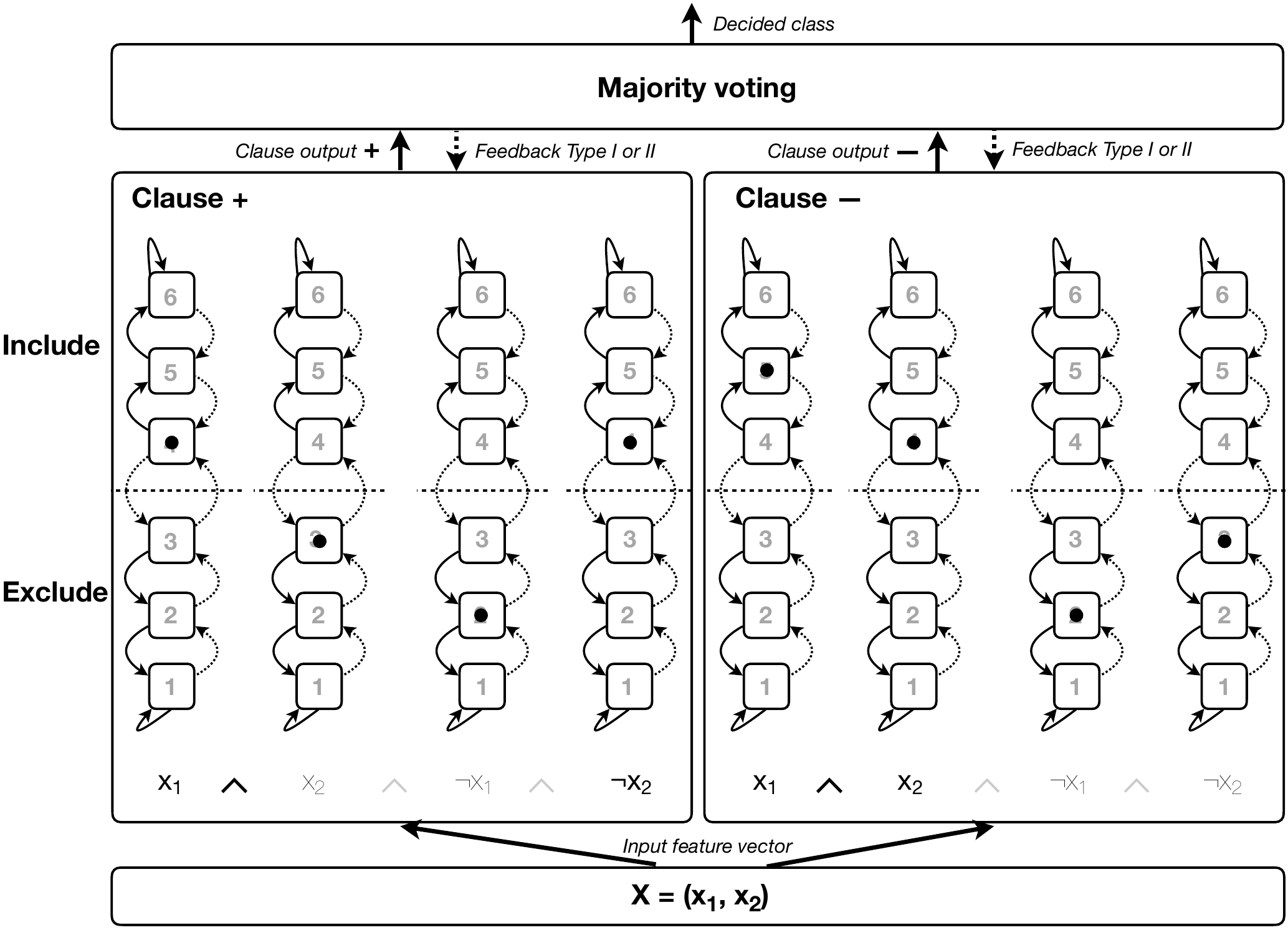}
\caption{The basic TM structure.}\label{figure:architecture_basic}
\end{figure}

The TM solves complex pattern recognition problems using \emph{conjunctive clauses} in propositional logic, composed by a collective of TAs. Its roots in game theory, the bandit problem, resource allocation, and frequent pattern mining are explored in depth in the original paper on the TM, which also includes pseudo code \cite{granmo2018tsetlin}. We here provide a more succinct overview, together with a running example to clarify core concepts.

\subsection{Structure}
The structure of the TM is shown in Figure~\ref{figure:architecture_basic}. As seen, the TM takes a vector $X$ of $o$ propositional variables as input: $\mathbf{X} = (x_k) \in \{0,1\}^{o}$, with $o$ being the size of the vector and $\{0,1\}^{o}$ its domain. The input vector in the figure, for instance, is of size $2$ and consists of the variables $x_1$ and $x_2$ with domain $\{(0,0), (0,1), (1,0), (1,1)\}$. 

The TM formulates patterns using \emph{conjunctive clauses}. A conjunctive clause is built from  \emph{literals}, that is, the input variables and their negations. With $o$ input variables, we have $2o$ literals, $\mathbf{L} = (l_k) \in \{0,1\}^{2o}$:
\begin{equation}
l_k =
\begin{cases}
x_k & \text{if } 1\le k \le o, \\
\lnot x_{k-o} & \text{otherwise }  (\text{i.e., } o+1\le k \le 2o).
\end{cases}
\end{equation}
Above, the first $o$ literals are the unnegated variables, while the $o$ following are the negated ones. Thus, in our example, there are four literals: $x_1, x_2, \lnot x_1$, and $\lnot x_2$. 

The number of clauses is a user set parameter that decides the expression power of the TM. Assume we have $m$ clauses in our TM structure. Each clause, denoted by $C_j$ and indexed by $j \in \{1, \ldots, m\}$, is simply a conjunction of a subset $\mathbf{I}_j \subseteq \{1, \ldots, 2o\}$ of the literals:
\begin{equation}\label{eq:clause}
C_j(\mathbf{X}) = \bigwedge_{k \in \mathbf{I}_j} l_k
\end{equation}
or equivalently:
\begin{equation}
C_j(\mathbf{X}) = \prod_{k \in \mathbf{I}_j} l_k.
\end{equation}
The subset $\mathbf{I}_j$ contains the indexes of the literals that have been \emph{Included} in the clause $j$. For the special case of $\mathbf{I}_j = \emptyset$, i.e., an empty clause, we have:
\begin{equation}
C_j(\mathbf{X}) = \left\{
	\begin{array}{ll}
		1  & \mathbf{during } \mbox{ learning} \\
		0 & \mathbf{otherwise}.
	\end{array}
\right.
\end{equation}
That is, during learning, empty clauses output $1$ and during classification they output $0$.
Accordingly, overall, the output $c_j = C_j(\mathbf{X}) \in \{0,1\}$ of a clause is fully specified by the input $\mathbf{X}$ and the selection of literals $\mathbf{I}_j$. Continuing our example, in the figure we have two clauses ($m=2$). The first consists of the literals with indexes $\mathbf{I}_1 = \{1, 4\}$: $C_1(\mathbf{X}) = x_1 \land \lnot x_2$. The second consists of the literals with indexes $\mathbf{I}_2 = \{1, 2\}$: $C_2(\mathbf{X}) = x_1 \land x_2$.

Each clause is further assigned a fixed polarity $p \in \{-1,+1\}$, which decides whether the clause output is negative or positive. In the figure, the first clause has positive polarity, whereas the second has negative. Positive clauses are used to recognize class $y=1$, while negative clauses are used to recognize class $y=0$. By default, half of the clauses are positive and half are negative. 

In the last step, the outputs of the clauses decide the class $\hat y\in\{0,1\}$ assigned to the input $\mathbf{X}$. That is, a summation operator aggregates the output of the clauses to form a majority vote. 

\subsection{Recognition}

We now look at recognition in more detail.  First of all, the output $c_j^p$ of each clause is organized as vectors $\mathbf{C}^+ = (c_j^+)\in \{0,1\}^\frac{m}{2}$ and $\mathbf{C}^- = (c_j^-)\in \{0,1\}^\frac{m}{2}$, grouped by polarity. Given an input $\mathbf{X}$, these outputs can then be calculated as follows (equivalent to Eqn. \ref{eq:clause}):
\begin{equation}\label{eq:clause_output}
c_j^p =
\begin{cases}
1 & \text{if } \mathbf{I}_j^p \subseteq \mathbf{I}_{L}^1, \\
0 & \text{otherwise}.
\end{cases}
\end{equation}
Here, the subset $\mathbf{I}_{L}^1 \subseteq \{1, \ldots, 2o\}$ contains the indexes of the literals that take the value $1$ for the current input $\mathbf{X}$: $\mathbf{I}_{L}^1 = \left\{k | l_k = 1,  1 \le k \le 2o \right\}$. To express this, the lower index $L$ stands for the literals and the upper index $1$ constrains these to those of value $1$. Thus, Eqn. \ref{eq:clause_output} compactly states that a clause outputs $1$ if and only if all of its literals are of value $1$. In Figure \ref{figure:architecture_basic}, for instance, the input $(x_1, x_2) = (1, 0)$ produces the literal values $l_1 = 1$, $l_2 = 0$, $l_3 = 0$, and $l_4 = 1$. Since the first clause only consists of $l_1$ and $l_4$, it outputs $1$. The second clause, on the other hand, consists of $l_1$ and $l_2$, hence it outputs $0$.

The final classification decision is made as follows. A majority vote is first organized by summing the output $\mathbf{C}^+$  of the positive clauses, and subtracting the output $\mathbf{C}^-$ of the negative clauses: $v =  \sum_j c_j^+ - \sum_j c_j^-$. This majority vote decides the prediction of the TM: $\hat{y} = (0 \le v$), with a tie resolved in favor of $y=1$.\footnote{Multi-class pattern recognition problems are modelled by employing multiple instances of this structure, replacing the threshold operator with an \textit{argmax} operator \cite{granmo2018tsetlin}.} So, in our example, the input $(x_1, x_2) = (1, 0)$ gathers a single positive vote, and no negative votes. Consequently, the classification becomes $\hat{y}=1$.

\subsection{Tsetlin Automata Collective}

We now introduce the collective of TAs, which decides the composition of each clause. There are $2o$ TAs per clause $j$, one for each literal $l_k \in \{x_1, \ldots, x_o, \lnot x_1, \ldots, \lnot x_o\}$. Each individual TA has a state $a_{j,k}^-, a_{j,k}^+ \in \{1, \ldots, 2N\}$, and the states of all of the TAs are organized in two $\frac{m}{2} \times o$ matrices. The first matrix, $\mathbf{A}^+ = (a_{j,k}^+) \in \{1, \ldots, 2N\}^{\frac{m}{2} \times 2o}$, contains the states of the TAs belonging to the positive polarity clauses, whereas the matrix $\mathbf{A}^- = (a_{j,k}^-) \in \{1, \ldots, 2N\}^{\frac{m}{2} \times 2o}$ covers the negative polarity clauses. In Figure \ref{figure:architecture_basic}, there are eight TAs, one per literal for clause $1$ (positive polarity) and one per literal for clause $2$ (negative polarity). Each of these TAs has $6$ states: $a_{j,k}^p \in \{1,\ldots, 6\}$.

The states decide the actions taken by the TAs. A TA decides to exclude its literal if in states $a_{j,k}^p \le N$, otherwise, it includes the literal. That is, for each clause $j$, some literals are \emph{Included}: $\mathbf{I}_j^p = \left\{ k | a_{j,k}^p > N,  1 \le k \le 2o\right\}$, while the remaining are \emph{Excluded}: $\mathbf{I}{_j^p}^c = \{1, \ldots, 2o\} \setminus \mathbf{I}_j^p$ (the complement of the included ones). So, in the figure, the TA associated with literal $x_1$ in the first clause is in state $4$ and thus $x_1$ is included in that clause. The second TA, on the other hand, is in state $3$ and, accordingly, literal $x_2$ is excluded. By checking the state of each TA in this manner, the composition of all of the clauses is decided.

\subsection{Learning}

Learning in the TM is based on coordinating the collective of TAs using a novel FSLA game. The game leverages resource-allocation \cite{Granmo2007d}
and frequent pattern mining \cite{Haugland2014}
principles, indicated by the feedback loop in Figure~\ref{figure:architecture_basic}. The feedback is handed out based on training examples $(\mathbf{X}, y)$, consisting of an input $ \mathbf{X}$ and an output $y$.

As explored below, the TM employs two kinds of feedback: Type~I and Type II. Type~I feedback jointly combats false negatives and overfitting by stimulating recognition of frequent patterns. Type II feedback, on the other hand, suppresses false positives by increasing the discrimination power of the patterns learnt.

Feedback is further regulated by the sum of clause outputs $v$ (the majority vote) and a target value $T \in \mathbb{Z}^>$ set for $v$ by the user. A larger $T$ (with a corresponding increase in the number of clauses) makes the learning more robust. This is because more clauses are involved in learning each specific pattern, introducing an ensemble effect. However, note that the resulting gain in accuracy comes at the cost of increased computational cost (cf. Table \ref{table:clauses_vs_accuracy_and_execution_time}).

\begin{figure}[ht]
\centering
\includegraphics[width=0.7\textwidth]{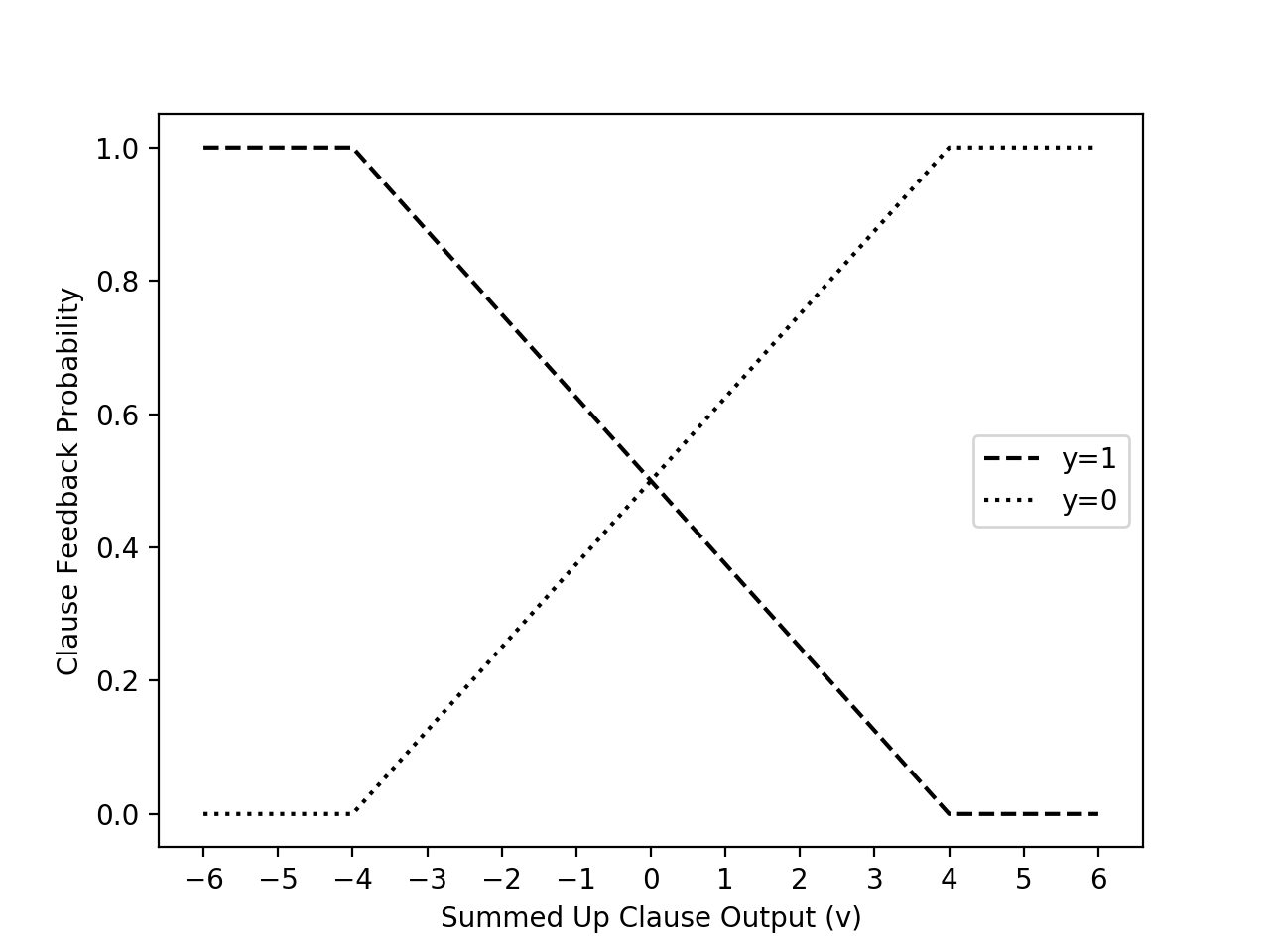}
\caption{Clause feedback probability for $T=4$.}\label{figure:clause_activation_probability}
\end{figure}

In the following, we first define the Type I and Type II feedback operations for clauses with positive polarity. As opposed to the feedback tables in \cite{granmo2018tsetlin}, the representation here is in a more compact matrix form.
\begin{itemize}
\item[Type I:] For training output $y=1$, TAs belonging to positive clauses are given Type I feedback to make $v$ approach $T$. A vector $\mathbf{R}_1^+ = (r_{j,1}^+) \in \{0,1\}^{\frac{m}{2}}$ picks out the positive clauses selected for feedback (the lower index of $\mathbf{R}_1^+$ refers to the output $y=1$):
\begin{equation}
r_{j,1}^+ =
\begin{cases}
1 & \text{with probability } \frac{T - \mathbf{clamp}(v, -T, T)}{2T}, \\
0 & \text{otherwise}.
\end{cases}
\end{equation}
Here, the clamp operation restricts $v$ to lie between $-T$ and $T$. Further, the lower index of $r_{j,1}^+$ refers to the clause $j$ and the output $y=1$. As illustrated by plot $y=1$ in Figure \ref{figure:clause_activation_probability}, clauses are randomly selected for feedback, with a larger chance of being selected for lower values of $v$. In effect, clauses are updated more aggressively the farther away the voting sum is from $T$, up to $-T$.

We now divide the Type I feedback into two parts, Type Ia and Type Ib. Type Ia reinforces \emph{Include} actions to make the patterns finer. Only TAs taking part in clauses that output $\True$ and whose associated literal takes the value $\True$ are select for Type Ia feedback. We collect the indexes of these TAs in the set $\mathbf{I}_{C^+}^{\text{Ia}} = \left\{(j,k) | l_{k} = 1 \land c_j^+ = 1 \land r_{j,1}^+ = 1 \right\}$.

Type Ib feedback is designed to reinforce \emph{Exclude} actions to combat over-fitting. Type Ib feedback is handed out to the TAs stochastically, using a user set parameter $s \ge 1.0$ (a larger $s$ provides finer patterns). The stochastic part of the calculation is organized in a matrix $\mathbf{Q}^+ = (q_{j,k}^+) \in \{0,1\}^{\frac{m}{2}\times2o}$ with entries:
\begin{equation}
q_{j,k}^+ =
\begin{cases}
1 & \text{with probability } \frac{1}{s},\\
0 & \text{otherwise}.
\end{cases}
\end{equation}
Here, the lower index of $q_{j,k}^+$ refers to the clause $j$ and the TA $k$, considering the positive clauses (upper index). The indexes of the TAs selected for Type Ib feedback are collected in the set $\mathbf{I}_{C^+}^{\text{Ib}} = \left\{(j,k) | (l_{k} = 0 \lor c_j^+ = 0) \land r_{j,1}^+ = 1 \land q_{j,k}^+ = 1 \right\}$.
That is, TAs taking part in clauses that output $0$, or whose associated literal takes the value 0, are selected for Type Ib feedback, however, only if stochastically pinpointed by $r_{j,1}^+ = 1$ and $q_{j,k}^+ = 1$. 

After the TAs has been selected for feedback, their states are updated by two state update operators $\oplus$ and $\ominus$:
$\mathbf{A}^+ \leftarrow \left( \mathbf{A}^+ \oplus {\mathbf{I}_{C^+}^{\text{Ia}}}\right) \ominus {\mathbf{I}_{C^+}^{\text{Ib}}}$
These operators add/subtract $1$ from the states of the singled out TAs, however, not beyond the given state space. In our example, if the TA associated with literal $x_1$ in the first clause is singled out for Type Ia feedback, it would move from state $4$ to state $5$, reinforcing the \emph{Include} action. If it instead receives Type Ib feedback, it would move to state $3$, reinforcing \emph{Exclude}. 

\item[Type II:] For training output $y=0$, TAs belonging to positive clauses are given Type II feedback to suppress clause output $1$. This, together with the negative clauses, makes $v$ approach $-T$. A matrix $\mathbf{R}_0^+ = (r_{j,0}^+) \in \{0,1\}^{\frac{m}{2}}$ picks out the positive clauses selected for feedback:
\begin{equation}
r_{j,0}^+ =
\begin{cases}
1 & \text{with probability } \frac{T + \mathbf{clamp}(v, -T, T)}{2T}, \\
0 & \text{otherwise}.
\end{cases}
\end{equation}
The lower index of $r_{j,0}^+$ refers to the clause $j$ and the output $y=0$, respectively. Here too, clauses are randomly selected for feedback, but now with a larger chance of being selected for higher $v$ (cf. plot $y=0$ in Figure \ref{figure:clause_activation_probability}). Next, the TAs selected are the ones that will turn clauses that output $1$ into clauses that output $0$, that is, those that have excluded literals that take the value $0$:
$\mathbf{I}_{C^+}^{\text{II}} = \left\{(j,k) | l_{k} = 0 \land c_j^+ = 1 \land r_{j,0}^+ = 1 \right\}$.
Again, the states of the selected TAs are updated using the dedicated operators:
$\mathbf{A}^+ \leftarrow \mathbf{A}^+ \oplus \mathbf{I}_{C^+}^{\text{II}}$. Thus, if the TA associated with literal $x_2$ in the first clause of our example is selected for Type II feedback, it would change state from $3$ to $4$, reinforcing the \emph{Include} action.
\end{itemize}

All of the above operations are for positive clauses. For negative clauses, Type I feedback is simply replaced with Type II feedback and vice versa!

\begin{figure}[!h]
\centering
\begin{subfigure}[t]{.49\textwidth}
\includegraphics[width=\linewidth]{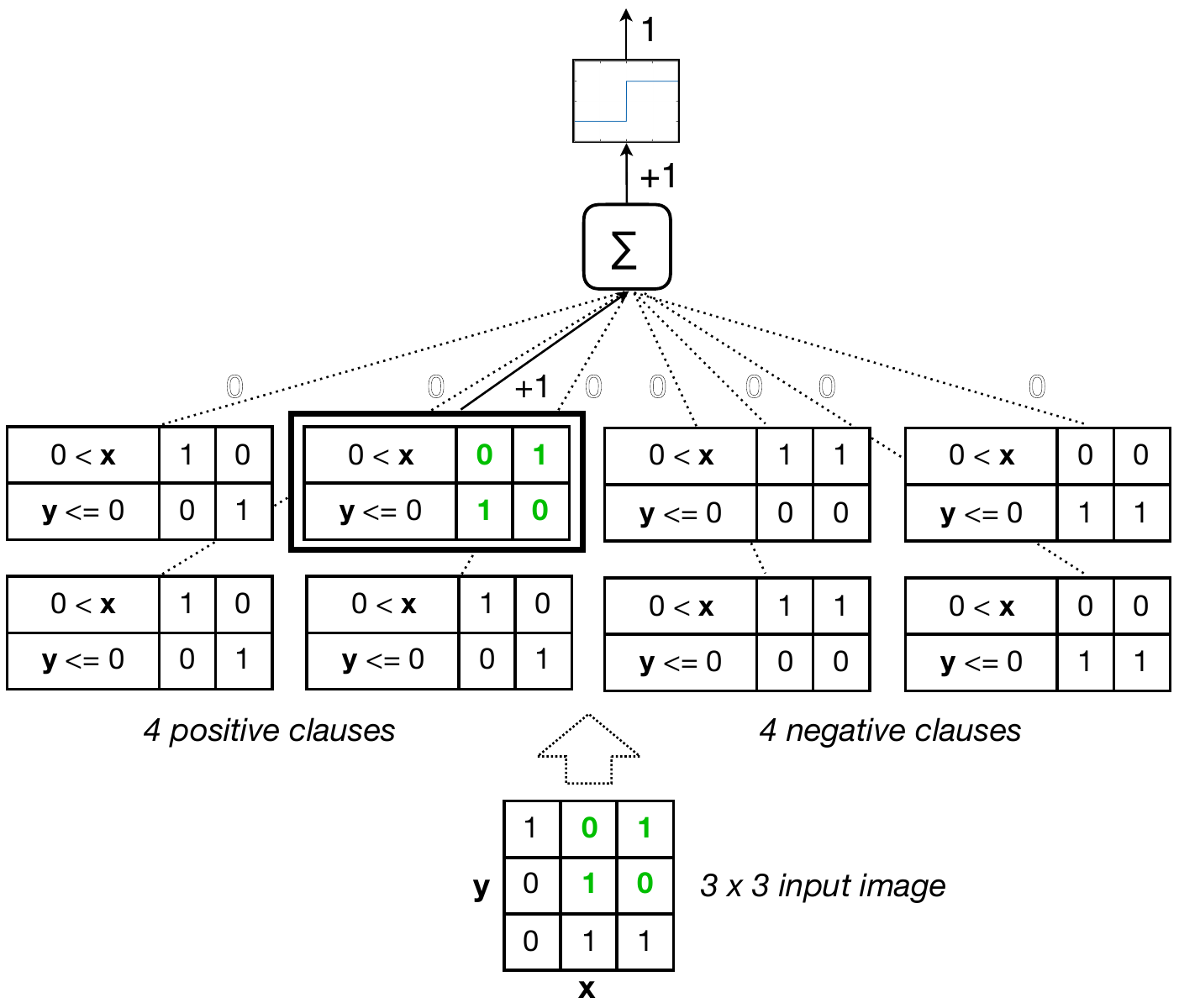}
\caption{}
\label{figure:inference_example}
\end{subfigure}
\hspace{0mm}
\begin{subfigure}[t]{.49\textwidth}
\includegraphics[width=\linewidth]{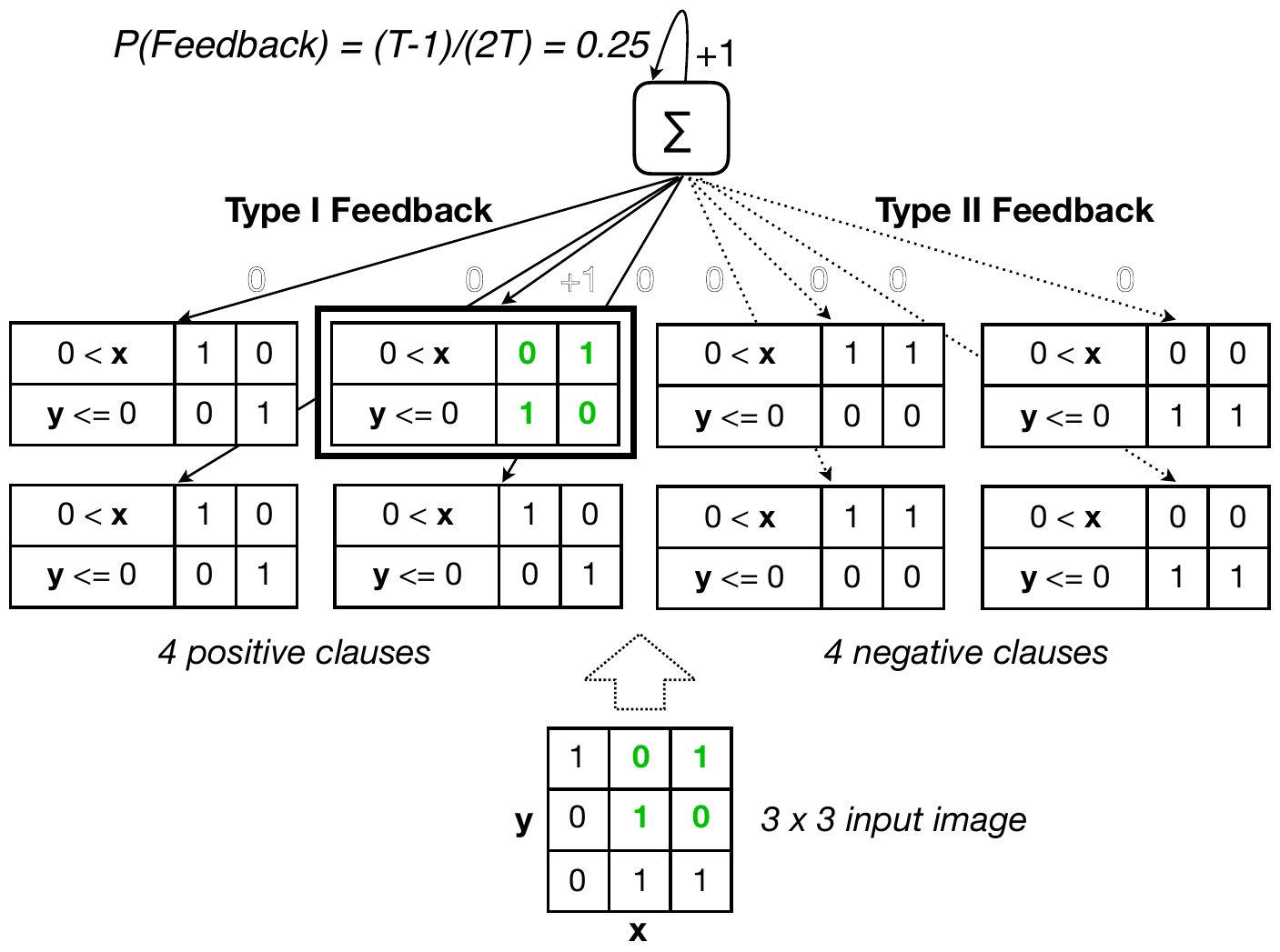}
\caption{}
\label{figure:learning_example}
\end{subfigure}
\caption{Example of inference (a) and learning (b) for the Noisy 2D XOR Problem.}
\end{figure}

\begin{figure}[!h]
\centering
\begin{subfigure}[t]{.49\textwidth}
\includegraphics[width=\linewidth]{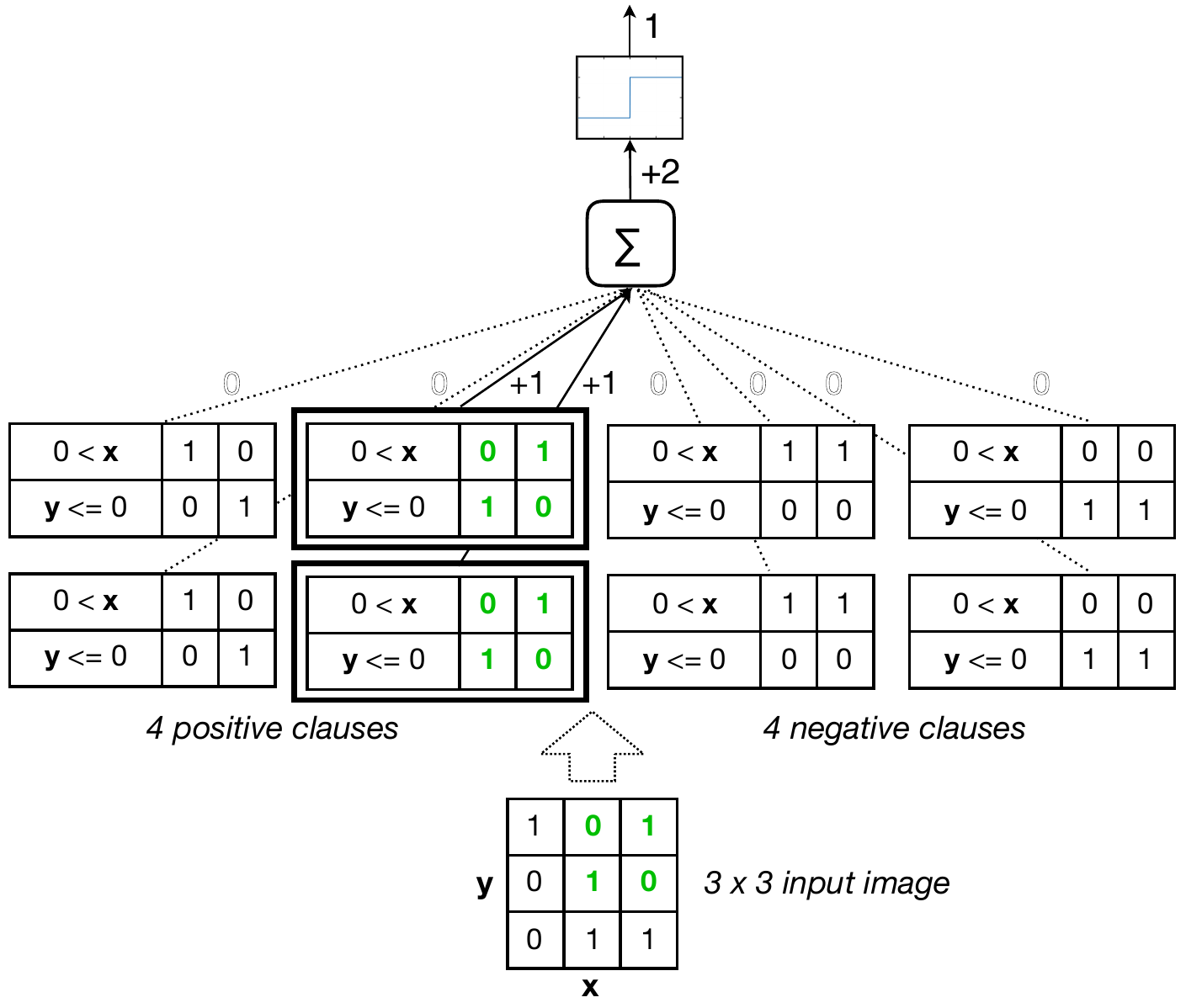}
\caption{}
\label{figure:goal_state}
\end{subfigure}
\hspace{0mm}
\begin{subfigure}[t]{.4\textwidth}
\includegraphics[width=\linewidth]{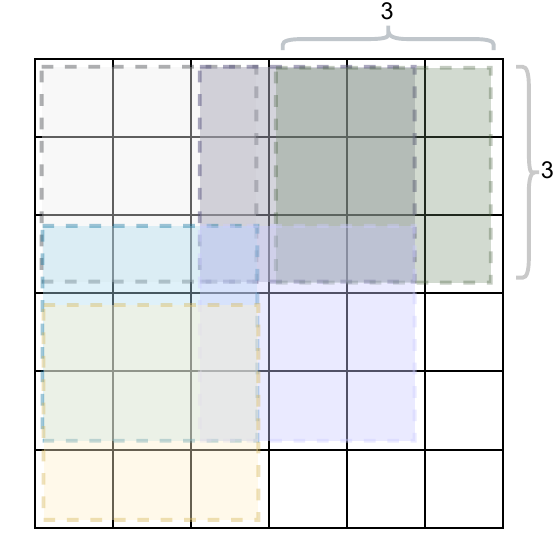}
\caption{}
\label{figure:convolution}
\end{subfigure}
\caption{(a) Goal state for the Noisy 2D XOR Problem. (b) Illustration of image, filter and patches.}\label{figure:example}
\end{figure}

\subsection{Integer Clause Weighting}

It turns out that introducing clause weighting improves TM performance, both accuracy- and computation-wise \cite{phoulady2019weighted}. By weighting the output of the clauses, one clause can replace multiple and the impact of a clause can be fine-tuned. In all brevity, classification is then performed as a weighted majority vote:

\begin{align}
s'(\mathbf x)=\sum_{j=1}^{m/2}w^+_j C^+_j(\mathbf x)-\sum_{j=1}^{m/2}w^-_jC^-_j(\mathbf x).
\end{align}

Whereas real-valued weighting was introduced in \cite{phoulady2019weighted}, we here address learning of integer weights, based on Stochastic Searching on the Line \cite{Oommen1997}:

\paragraph{Initialization.} First of all, the weights are initialized to $1$:
\begin{align}
w^+_j&\gets 1,\\
w^-_j&\gets 1.
\end{align}
In other words, the behaviour of the clauses are identical to those of a standard TM at the start of learning.

\paragraph{Weight Updating.} The TA state updating is unmodified with weights. The weights, in turn, are updated based on Type~I and Type~II feedback.
\begin{itemize}
\item[Type I:] For Type~I feedback, a clause weight is increased by $1$, however, only if the clause outputted $1$:
\begin{align}
w^+_j&\gets w^+_j + 1, \quad\text{if }C^+_j(\mathbf x)=1,\\
w^-_j&\gets w^-_j + 1,\quad\text{if }C^-_j(\mathbf x)=1.
\end{align}
\item[Type II:] For Type~II feedback, we instead update the weights by subtracting $1$, with $1$ being the minimum weight value:
\begin{align}
w^+_j&\gets w^+_j - 1,\quad\text{if }C^+_j(\mathbf x)=1 \land w^+_j > 1,\\
w^-_j&\gets w^-_j - 1,\quad\text{if }C^-_j(\mathbf x)=1 \land w^-_j > 1.
\end{align}
\end{itemize}
When a clause evaluates to $0$, its weight remains unchanged:
\begin{align}
w^+_j&\gets w^+_j,\quad\text{if }C^+_j(\mathbf x)=0,\\
w^-_j&\gets w^-_j,\quad\text{if }C^-_j(\mathbf x)=0.
\end{align}
As seen, for Type~I feedback, the weights are increased to strengthen the impact of the associated clauses, thus reinforcing true positive frequent patterns. For Type~II feedback, on the other hand, the weights are decreased instead. This is to diminish the impact of the associated clauses, thus combating false positives, increasing discrimination power.

\section{The Convolutional Tsetlin Machine}
\label{sec:CTM}

Consider a set of images $\mathcal{X}=\{\mathbf{X}_e | 1 \le e \le E\}$, where $e$ is the index of the images. Each image is of size $X \times Y$, and consists of $Z$ binary layers (which together encode the pixel colors using thresholding \cite{Abeyrathna2019}, one-hot encoding, or any other appropriate binary encoding). A classic TM models such an image with an input vector $\mathbf{X} = (x_k) \in \{0, 1\}^{X\times Y \times Z}$ that contains $X\times Y \times Z$ propositional variables. Further, each clause is composed from $X\times Y \times Z \times 2$ literals. Inspired by the impact convolution has had on deep neural networks, we here introduce the Convolutional Tsetlin Machine (CTM).

\textbf{Interpretable Rule-based Filters.} The CTM uses filters with spatial dimensions $W\times W$, again with $Z$ binary layers. Further, the clauses of the CTM take the role of filters. Each clause is accordingly composed from $W\times W \times Z \times 2$ literals. Additionally, to make the clauses location-aware \cite{Liu2018}, we augment each clause with binary encoded coordinates. Location awareness may prove useful in applications where both patterns and their location are distinguishing features, e.g. recognition of facial features such as eyes, eyebrows, nose, mouth, etc. in facial expression recognition. In all brevity, when applying a filter of size $W \times W$ on an image $(x_k) \in \{0, 1\}^{X\times Y \times Z}$, the filter will be evaluated on $B = B_X \times B_Y$ image patches. Here, $B_X = \ceil*{\frac{X-W}{d}}+1$ and $B_Y = \ceil*{\frac{Y-W}{d}}+1$,  with $d$ being the step size of the convolution. Each image patch thus has a certain location within the image, and we augment the input vector with the coordinates of this location. We denote the resulting augmented input vector $\mathbf{X}^b$:  $\mathbf{X}^b = (x_k) \in \{0, 1\}^{W \times W \times Z + B_X + B_Y}$. As seen, the input vector is extended with one propositional variable per position along each dimension, with the position being encoded using thresholding \cite{Abeyrathna2019} or one-hot encoding. Figure \ref{figure:convolution} illustrates an example of the image, patches, and a filter for $X=Y=6$, $W=3$, and $d=2$. In this example, the 3$\times$3 filter moves from left to right, from top to bottom,  $2$ pixels per step.

\textbf{Recognition.} The CTM uses the classic TM procedure for recognition (see Sect. \ref{sec:TM}). However, for the CTM each clause outputs $B$ values per image (one value per patch), as opposed to a single output for the TM (Eq. \ref{eq:clause_output}). We denote the output of a positive clause $j$ on patch $b$ by $c_j^{b,+}$.  To turn the multiple outputs $c_j^{1,+}, \ldots, c_j^{B,+}$
of clause $j$ into a single output denoted by $c_j^+$, we simply OR the individual outputs:
\begin{equation}
c_j^+ = \bigvee_{b=1}^B c_j^{b,+}. 
\end{equation}

\textbf{Learning.} Learning in the CTM leverages the TM learning procedure. As seen in Sect. \ref{sec:TM}, Type Ia, Type Ib, and Type II feedback are influencing each clause $j$ based on the literals of the input vector $\mathbf{X}$. For the CTM, the input vector is an image patch, and there are $B$ patches in an image. There is thus $B$ literal inputs $\mathbf{L}^b$, $1 \le b \le B$, per clause. Therefore, to decide which patch to use when updating a clause, the CTM randomly  selects a single patch among the patches that made the clause evaluate to $1$. The clause is then updated according to this patch. That is, the input $\mathbf{X}^b$ is drawn from the set: $\{\mathbf{X}^b | c_j^{b,+} = 1, 1 \le b \le B\}$. Observe that if the set is empty, only Type Ib feedback is applicable, and then the input vector is not needed. For non-empty sets, the TAs to be updated are finally singled out using the randomly selected patch:
\begin{align}
&\mathbf{I}_{C^+}^{\text{Ia}} = \left\{(j,k) | l_{k}^b = 1 \land c_j^+ = 1 \land r_{j,1}^+ = 1 \right\},\\
&\mathbf{I}_{C^+}^{\text{Ib}} = \left\{(j,k) | (l_{k}^b = 0 \lor c_j^+ = 0) \land r_{j,1}^+ = 1 \land q_{j,k}^+ = 1 \right\},\\
&\mathbf{I}_{C^+}^{\text{II}} = \left\{(j,k) | l_{k}^b = 0 \land c_j^+ = 1 \land r_{j,0}^+ = 1 \right\}.
\end{align}
The reason for randomly selecting a patch is to have each clause extract a certain sub-pattern, and the randomness of the uniform distribution statistically spreads the clauses for different sub-patterns in the target image.  Finally, observe that the computational complexity of the CTM grows linearly with the number of clauses $m$, and with the number of patches $B$. However, the computations can be easily parallelized due to their decentralized nature.

\textbf{Step-by-step Walk-through of Inference on Noisy 2D XOR.} Rather than providing hand-crafted features which can be used for image classification, the CTM learns feature detectors. We will explain the workings of the CTM by an illustrative example of noisy 2D XOR recognition and learning (see Figure \ref{figure:example} and Sect. \ref{sec:empirical_results}). Consider the CTM depicted in Figure \ref{figure:inference_example}. It consists of four positive clauses which represent XOR patterns that must be present in a positive example image (positive features) and four negative clauses which represent patterns that will not trigger a positive image classification (negative features). The number of positive and negative clauses is a user-defined parameter. The bit patterns inside each clause are represented by the output of eight TAs, two for each bit in a $2\times2$ filter.

Consider the $3\times3$ image shown in Figure \ref{figure:learning_example}. The filter represented by the second positive clause matches the patch in the top-right corner of the image and it is the only clause with output $1$; similarly, none of the negative clauses respond since their patterns do not match the pattern found in the current patch (Figure \ref{figure:learning_example}). Thus, the TM's combined output is $v=1$. Learning of feature detectors proceeds as follows: with the CTM’s voting target set to $T=2$, the probability of feedback is $\frac{T-v}{2T} = \frac{1}{4}$, and thus learning takes place, which pushes the CTM’s output $v$ towards $T=2$. Note that Type I feedback reinforces true positive output and reduces false negative output whereas Type II feedback reduces false positive output.

A subsequent state of the CTM is shown in Figure \ref{figure:goal_state}. There are now two positive clauses which detect their pattern in the top-right corner patch. The combined output of all clauses is $2$; thus, no further learning is necessary for the detection of the XOR pattern in this patch. Also, the location of the occurrence of each pattern is included. The location information uses a bit representation as follows: Suppose an XOR pattern occurs at the three X-coordinates $1$, $4$, and $6$. For the corresponding binary location representation, these coordinates are considered as thresholds: If a coordinate is greater than a threshold, then the corresponding bit in the binary representation will be $0$; otherwise, it is set to $1$. Thus, the representation of the X-coordinates $1$, $4$, and $6$ will be ``111", ``011" and ``001", respectively. These representations of the location of $2 \times 2$ patterns are also learned by TAs.

\begin{table*}[!h]
\centering
\begin{tabular}{ c|c|c|c|c|c } 
\hline
&Search Range& 2D Noisy XOR & MNIST & K-MNIST & Fashion-MNIST\\
\hline
\hline
\#Class Clauses&$1-8~000$&$40$&$8~000$&$8~000$&$8~000$\\
T&$1-20~000$&$60$&$10~000$&$10~000$&$10~000$\\
s&$1.0-20.0$&$3.9$&$5.0$&$10.0$&$10.0$\\
W&$2-20$&$2$&$10$&$10$&$10$\\
\big|Z\big|&$1$&$1$&$1$&$1$&$1$\\
Clause Weighting&Yes/No&No&Yes&Yes&Yes\\
\hline
\end{tabular}
\caption{CTM configurations.}\label{table:configuration}
\end{table*}

\begin{table*}[!!h]
    \centering
    \begin{tabular}{ccccc}
    \textbf{0: U L} &  \textbf{0: U R} &  \textbf{9: U L} &  \textbf{9: U R} &  \textbf{4: L L}\\
    \begin{tabular}{c}
    00*000*000\\
    *000*0*000\\
    **00*00000\\
    00000000**\\
    00*000****\\
    000000***1\\
    000**0*111\\
    0*00**111*\\
    *****111*0\\
    00**11**00\\
    \hline
    0 $<$ X $\le$ 6\\
    5 $<$ Y $\le$ 8
    \end{tabular}
    &
    \begin{tabular}{c}
    0*********\\
    0******0**\\
    ********00\\
    *111111**0\\
    *111*11**0\\
    *1*1*11***\\
    111**11**0\\
    1****111**\\
    11***111**\\
    *****111**\\
    \hline
    8 $<$ X $\le$ 17\\
    0 $<$ Y $\le$ 6
    \end{tabular}
    &
    \begin{tabular}{c}
    0*0*****1*\\
    00000**1**\\
    000*******\\
    00***1**00\\
    000**1**00\\
    00**1*00*0\\
    00****00**\\
    0*******11\\
    0*********\\
    000*******\\
    \hline
    2 $<$ X $\le$ 7\\
    5 $<$ Y $\le$ 10
    \end{tabular}
    &
    \begin{tabular}{c}
    0**0*****0\\
    *00*0*****\\
    **0*******\\
    *******0**\\
    *11****0*0\\
    ****0**000\\
    ****1**00*\\
    ****1***0*\\
    ****11**00\\
    ********0*\\
    \hline
    12 $<$ X $\le$ 18\\
    2 $<$ Y $\le$ 8
    \end{tabular}
    &
    \begin{tabular}{c}
    ******1***\\
    *11*1*1***\\
    ***111****\\
    *****11***\\
    *000**1***\\
    000**1****\\
    0000*1*0**\\
    0*00*1**0*\\
    0**0***0**\\
    0000****0*\\
    \hline
    7 $<$ X $\le$ 11\\
    12 $<$ Y $\le$ 17
    \end{tabular}\\
    \textbf{0: L L} & \textbf{0: L R} &  \textbf{9: L L} &  \textbf{9: L R} &  \textbf{3: U R}\\
    \begin{tabular}{c}
    *000**1***\\
    **********\\
    **0******0\\
    *00****110\\
    *********0\\
    *0*0***1*0\\
    ******11**\\
    ***0******\\
    **000**1*1\\
    0*0****11*\\
    \hline
    0 $<$ X $\le$ 5\\
    10 $<$ Y $\le$ 16
    \end{tabular}
    &
    \begin{tabular}{c}
    *******1**\\
    ******1***\\
    ******1***\\
    *****1**1*\\
    ****1*1***\\
    ****11****\\
    *1*11*1***\\
    1**11*****\\
    *11*11****\\
    *1*1******\\
    \hline
    4 $<$ X $\le$ 18\\
    13$<$ Y $\le$ 18
    \end{tabular}
    &
    \begin{tabular}{c}
    0000000***\\
    00000001**\\
    0000000110\\
    0000000*1*\\
    000000*0*1\\
    00000000**\\
    0000000000\\
    0000*00000\\
    0000000000\\
    0000000000\\
    \hline
    0 $<$ X $\le$ 3\\
    8 $<$ Y $\le$ 14
    \end{tabular}
    &
    \begin{tabular}{c}
    ******00**\\
    *****000*0\\
    **11*0000*\\
    **11*00000\\
    1*11000*00\\
    11**0000*0\\
    111****00*\\
    1**0000000\\
    1**00*00**\\
    1*0**00*00\\
    \hline
    12$<$ X $\le$ 16\\
    9 $<$ Y $\le$ 15
    \end{tabular}
    &
    \begin{tabular}{c}
    **0000*000\\
    00000*0000\\
    ***0000000\\
    *11**00000\\
    *1*1**0000\\
    0**1*00*00\\
    0*1**0***0\\
    **1*0000*0\\
    *1***0*000\\
    ***0000000\\
    \hline
    12 $<$ X $\le$ 18\\
    0$<$ Y $\le$ 3
    \end{tabular}
    \end{tabular}
    \caption{Example 10$\times$10 bit patterns produced by CTM for MNIST, including valid convolution positions. Here ``0: UL" means ``upper left" of the image for digit ``0". We can see clearly that ``0: UL", ``0: UR", ``0: LL" and ``0: LR" jointly construct the shape ``0''. The clauses for the other digits behave similarly, and we thus just illustrate selected patch positions for each digit.}\label{table:example_patterns}
    \label{tab:bit_pattern}
\end{table*}

\section{Empirical Results}
\label{sec:empirical_results}
In this section, we evaluate the CTM on four different datasets.

\textbf{2D Noisy XOR.} The 2D Noisy XOR dataset contains $4 \times 4$ binary images, $2500$ training examples and $10~000$ test examples. The image bits have been set randomly, except for the $2 \times 2$ patch in the upper right corner, which reveals the class of the image. A diagonal line is associated with class $1$, while a horizontal or vertical line is associated with class $0$.
Thus the dataset models a 2D version of the XOR-relation. Furthermore, the dataset contains a large number of random non-informative features to measure susceptibility towards the curse of dimensionality. To examine robustness towards noise we have further randomly inverted $40\%$ of the outputs in the training data.

\textbf{MNIST.} The MNIST dataset  has been used extensively to benchmark machine learning algorithms, consisting of $28\times 28$ grey scale images of handwritten digits~\cite{lecun1998gradient}.

\textbf{Kuzushiji-MNIST.} This dataset contains $28 \times 28$ grayscale images of Kuzushiji characters, cursive Japanese. Kuzushiji-MNIST is more challenging than MNIST because there are multiple distinct ways to write some of the characters  \cite{Clanuwat2018}.

\textbf{Fashion-MNIST.} This dataset contains $28 \times 28$ grayscale images of articles from the Zalando catalogue, such as t-shirts, sandals, and pullovers  \cite{xiao2017}. This dataset is quite challenging, with a human accuracy of $83.50$\%.

The latter three datasets contain $60~000$ training examples and $10~000$ test examples. We binarize these datasets using an adaptive Gaussian thresholding procedure with window size $11$ and threshold value $2$. Accordingly, the CTM operates on images with merely 1 bit per pixel. Table \ref{table:results} reports test accuracy for the CTM, while Table \ref{table:configuration} contains the corresponding configurations. The results are based on single-run averages, obtained from the last $100$ epochs of $250$. All of the experiments were run on a NVIDIA DGX-2 server. Note that these datasets are rather simple to solve for deep CNNs, and have been selected to facilitate our study of interpretability.

\begin{table*}[!!h]
\centering
\begin{tabular}{c|c|c|c|c|c|c} 
\hline
\#Clauses per class&250&500&1000&2000&4000&8000\\
\hline
\hline
MNIST accuracy (\%)&98.82&98.98&99.14&99.22&99.28&99.33\\
MNIST execution time (s)&15.6&15.8&18.9&20.4&32.4&39.0\\
\hline
K-MNIST accuracy (\%)&92.75&93.86&94.89&95.40&95.85&96.08\\
K-MNIST execution time (s)&15.5&17.2&19.7&21.1&33.5&40.5\\
\hline
F-MNIST accuracy (\%)&88.25&88.79&89.42&89.89&90.65&91.18\\
F-MNIST execution time (s)&16.9&17.2&17.8&25.0&27.8&52.2\\
\hline
\end{tabular}
\caption{CTM mean test accuracy and execution time per epoch for an increasing number of clauses.}\label{table:clauses_vs_accuracy_and_execution_time}
\end{table*}

\begin{table*}[!!h]
\centering
\begin{tabular}{ c|c|c|c|c } 
\hline
Model & 2D N-XOR & MNIST & K-MNIST & F-MNIST\\
\hline
\hline
4-Nearest Neighbour \cite{Clanuwat2018,xiao2017}&$61.62$&$\mathit{97.14}$&$\mathit{91.56}$&$85.40$\\
SVM \cite{Clanuwat2018}&$94.63$&$\mathit{98.57}$&$\mathit{92.82}$&$\mathit{89.7}$\\
Random Forest \cite{LiangAAAI19}&70.73&$\mathit{97.3}$&-&$\mathit{81.6}$\\
Gradient Boosting Classifier \cite{xiao2017}&87.15&$\mathit{96.9}$&-&$\mathit{88.0}$\\
Simple CNN \cite{Clanuwat2018,LiangAAAI19}&$91.05$&$\mathit{99.06}$&$\mathit{95.12}$&$\mathit{90.7}$\\
BinaryConnect \cite{courbariaux2015binaryconnect}&-&$\mathit{98.99}$&-&-\\
FPGA-accelerated BNN \cite{Lammie2019}&-&$\mathit{98.70}$&-&-\\
Logistic Circuit (binary) \cite{LiangAAAI19}&-&$\mathit{97.4}$&-&$\mathit{87.6}$\\
Logistic Circuit (real-valued) \cite{LiangAAAI19}&-&$\mathit{99.4}$&-&$\mathit{91.3}$\\
PreActResNet-18 \cite{Clanuwat2018}&-&$\mathit{99.56}$&$\mathit{97.82}$&$\mathit{92.00}$\\
ResNet18 + VGG Ensemble \cite{Clanuwat2018}&-&$\mathit{99.60}$&$\mathit{98.90}$&-\\
TM&$99.12$&$98.57$&$92.03$&$90.09$\\
\hline
CTM (Mean)&$99.99 \pm 0.0$&$99.33 \pm 0.0$&$96.08 \pm 0.01$&$91.18 \pm 0.01$\\
CTM (95 \%ile)&$100.0$&$99.38$&$96.25$&$91.39$\\
CTM (Peak)&$100.0$&$99.40$&$96.31$&$91.50$\\
\hline
\end{tabular}
\caption{Empirical results - test accuracy in percent.}\label{table:results}
\end{table*}

\begin{table*}[!!h]
\centering
\begin{tabular}{ c|c|c } 
\hline
Model & MNIST & F-MNIST\\
\hline
\hline
Logistic Circuit (binary) \cite{LiangAAAI19}&$1~072$ kB&$2~456$ kB\\
Logistic Circuit (real-valued) \cite{LiangAAAI19}&$728$ kB&$1~768$ kB\\
CNN w/3 conv. layers \cite{LiangAAAI19}&$8~784$ kB&$8~784$ kB\\
ResNet \cite{LiangAAAI19}&$19~352$ kB&$19~352$ kB\\
TM w/$4~000$ class clauses&$7~657$ kB&$7~657$ kB\\
\hline
CTM w/$250$ class clauses&$127$ kB&$127$ kB\\
CTM w/$8~000$ class clauses&$2~109$ kB&$2~109$ kB\\
\hline
\end{tabular}
\caption{Model size in kilobytes (kB), assuming 32-bit real-valued weights.}\label{table:model_size}
\end{table*}

\begin{figure}[ht]
\centering
\pgfplotstableread{mnist_stats_8000_10000_5.00.dat}{\mnist}
\pgfplotstableread{kmnist_stats_8000_10000_10.00.dat}{\kmnist}
\pgfplotstableread{fmnist_stats_8000_10000_10.00.dat}{\fmnist}
\begin{tikzpicture}
\tikzstyle{more densely dashed}=[dash pattern=on 5pt off 1pt]
	\begin{axis}[
	    ymin=75, ymax=100,
	    xmin=0, xmax=250,
		xlabel=Epoch,
		ytick distance=2,
		ylabel=Accuracy (\%),
        width=0.8\textwidth,
        height=0.6\textwidth,
        legend pos=south east, legend cell align={left},
        ymajorgrids=true, grid style={dotted,gray}
	]
	\addplot [black, dashed] table {\mnist};
	\addplot [black, more densely dashed] table {\kmnist};
	\addplot [black, densely dotted] table {\fmnist};
	\addlegendentry{MNIST}
	\addlegendentry{K-MNIST}
	\addlegendentry{F-MNIST}
	\end{axis}
\end{tikzpicture}
\caption{Single-run test accuracy per epoch for CTM on MNIST, K-MNIST, and F-MNIST.}
\label{figure:accuracy_per_epoch}
\end{figure}
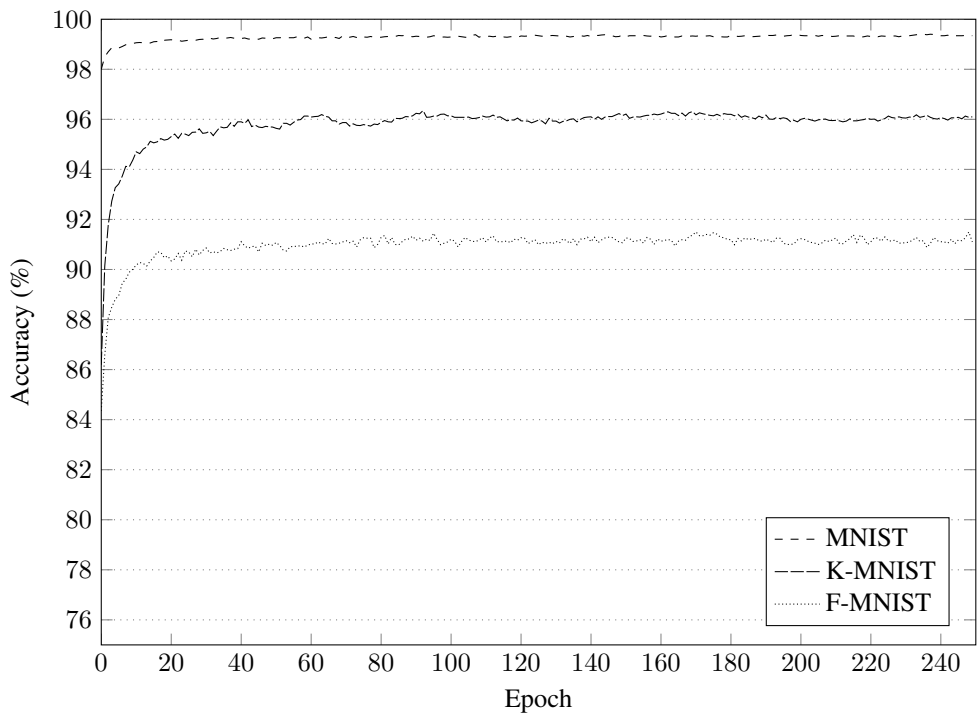

 The hyperparameters were set using a light manual binary search. One particularly critical hyperparameter is the number of clauses employed per class. Table \ref{table:clauses_vs_accuracy_and_execution_time} reports execution time per epoch as well as mean test accuracy for an increasing number of class clauses. As seen, test accuracy increases steadily with the number of clauses employed, while execution time increases sub-linearly due to parallelization.
 
  The CTM performs rather robustly, providing tight $95$\% confidence intervals for the mean performance. As seen in Table  \ref{table:results}, we have also included results for selected popular algorithms, as points of reference. Note that the CTM is an interpretable approach that does not employ multiple layers at this stage. Therefore, we believe a fair comparison should target traditional/simple CNNs. Results listed in italic are reported in the corresponding papers. Results for BinaryConnect and FPGA-accelerated BNNs on K-MNIST and Fashion-MNIST were not available, so are not reported. Notice that the CTM outperforms the binary CNNs for all the datasets, as well several simple baseline configurations, including 4-layer CNNs, Random Forests, Gradient Boosting, SVMs, and a 4-nearest neighbour classifier. Further observe that both the TM and the CTM obtain higher accuracy than Logistic Circuits \cite{LiangAAAI19} with binary data. Thus, in our experiments, the TM learning paradigm outperforms the probabilistic circuits paradigm \cite{LiangAAAI19,Peharzetal18} at extracting information from binary representations. However, only the CTM performs competitively with Logistic Circuits when the Logistic Circuits are given the advantage of real-valued inputs. Note that the CTM supports using multiple bits to encode the color of each pixel, however, in the present paper, we only investigate single bit representations. Finally, the CTM is outperformed by the more advanced deep learning network architectures PreActResNet-18 and ResNET18+VGG.
  
  Figure \ref{figure:accuracy_per_epoch} depicts test accuracy for the CTM on MNIST, K-MNIST, and F-MNIST, epoch-by-epoch. As seen, test accuracy climbs quickly in the first epochs, e.g., passing $99$\% already in epoch $8$ for MNIST. MNIST test accuracy peaks at $99.4$\% after $234$ epochs, while K-MNIST peaks at $96.31$\% after $169$ epochs. For F-MNIST, test accuracy surpasses $90.0$\% in epoch $9$ and peaks at $91.5$\% in epoch $170$.
  
In Table \ref{table:model_size}, the size of the trained models are listed for MNIST and F-MNIST. Size is measured in kilobytes of memory used. The memory usage of CTM is similar to Logistic Circuits, which both use significantly less memory than the neural network architectures. A potential advantage of CTM is the ability to trade off accuracy against computation time and memory usage, by reducing the number of clauses employed. E.g., a CTM with 250 clauses per class reduces memory usage by more than an order of magnitude at a relatively small loss in test accuracy (cf. Table \ref{table:clauses_vs_accuracy_and_execution_time}). 
 
Table \ref{table:example_patterns} contains example patterns produced by CTM for MNIST (with one bit per pixel using $8000$ clauses). The ``*" symbol can either take the value ``0" or ``1". The remaining bit values require strict matching. The examples have been selected to illustrate how several patterns together form complete numbers. As seen, the patterns are relatively easy to interpret for humans compared to, e.g., a neural network. They are also efficient to evaluate for computers, involving only logical operators, followed by summation.
 
\section{Conclusion and Further Work}
\label{sec:conclusion}

This paper introduced the Convolutional Tsetlin Machine (CTM), leveraging the learning mechanism of the Tsetlin Machine (TM). Whereas the TM categorizes images by employing each clause once per $X \times Y$ input, the CTM uses each clause as a  $W \times W$ convolution filter. The filters learned by the CTM are interpretable, being formulated using propositional formulae. To make the clauses location-aware, each patch is further enhanced with its coordinates within the image. Location awareness may prove useful in applications where both patterns and their location are distinguishing features. By \emph{randomly} selecting which patch to learn from, the standard Type I and Type II feedback of the classic TM can be employed directly. In this manner, the CTM obtains results on MNIST, Kuzushiji-MNIST, Fashion-MNIST, and the 2D Noisy XOR Problem that compare favorably with simple 4-layer CNNs, Logistic Circuits, as well as two binary neural network architectures.

In our further work, we intend to investigate more advanced binary encoding schemes, to go beyond grey-scale images (e.g., addressing CIFAR-10 and ImageNet). We further intend to develop schemes for deeper CTMs, with the first step being a two-layer CTM, to introduce more compact and expressive patterns with nested propositional formulae.

\bibliographystyle{abbrv}
\bibliography{references}

\begin{thebibliography}{10}

\bibitem{abeyrathna2019nonlinear}
K.~D. {Abeyrathna}, O.-C. {Granmo}, X.~{Zhang}, L.~{Jiao}, and M.~{Goodwin}.
\newblock {The Regression Tsetlin Machine - A Novel Approach to Interpretable
  Non-Linear Regression}.
\newblock {\em Philosophical Transactions of the Royal Society A}, 378, 2019.

\bibitem{Agrawal1993}
R.~Agrawal, T.~Imieli\'{n}ski, and A.~Swami.
\newblock Mining association rules between sets of items in large databases.
\newblock {\em SIGMOD Rec.}, 22(2):207--216, 1993.

\bibitem{berge2019}
G.~T. {Berge}, O.-C. {Granmo}, T.~O. {Tveit}, M.~{Goodwin}, L.~{Jiao}, and
  B.~V. {Matheussen}.
\newblock {U}sing the {T}setlin {M}achine to {L}earn {H}uman-{I}nterpretable
  {R}ules for {H}igh-{A}ccuracy {T}ext {C}ategorization with {M}edical
  {A}pplications.
\newblock {\em IEEE Access}, 7:115134--115146, 2019.

\bibitem{Bouhmala2010}
N.~Bouhmala and O.-C. Granmo.
\newblock {Stochastic Learning for SAT-Encoded Graph Coloring Problems}.
\newblock {\em International Journal of Applied Metaheuristic Computing},
  1(3):1--19, 2010.

\bibitem{Clanuwat2018}
T.~{Clanuwat}, M.~{Bober-Irizar}, A.~{Kitamoto}, A.~{Lamb}, K.~{Yamamoto}, and
  D.~{Ha}.
\newblock {Deep Learning for Classical Japanese Literature}.
\newblock {\em arXiv:1812.01718}, Dec 2018.

\bibitem{courbariaux2015binaryconnect}
M.~Courbariaux, Y.~Bengio, and J.-P. David.
\newblock Binaryconnect: Training deep neural networks with binary weights
  during propagations.
\newblock In {\em Advances in neural information processing systems}, pages
  3123--3131, 2015.

\bibitem{Darabi2018}
S.~{Darabi}, M.~{Belbahri}, M.~{Courbariaux}, and V.~{Partovi Nia}.
\newblock {BNN+: Improved Binary Network Training}.
\newblock {\em arXiv:1812.11800}, Dec 2018.

\bibitem{Abeyrathna2019}
K.~{Darshana Abeyrathna}, O.-C. {Granmo}, X.~{Zhang}, and M.~{Goodwin}.
\newblock {A Scheme for Continuous Input to the Tsetlin Machine with
  Applications to Forecasting Disease Outbreaks}.
\newblock {\em arXiv:1905.04199}, May 2019.

\bibitem{feldman9}
V.~Feldman.
\newblock {H}ardness of {A}pproximate {T}wo-{L}evel {L}ogic {M}inimization and
  {PAC} {L}earning with {M}embership {Q}ueries.
\newblock {\em Journal of Computer and System Sciences}, 75(1):13--26, 2009.

\bibitem{Ghavipour2018}
M.~Ghavipour and M.~R. Meybodi.
\newblock {A streaming sampling algorithm for social activity networks using
  fixed structure learning automata}.
\newblock {\em Applied Intelligence}, 2018.

\bibitem{Gittins1979}
J.~Gittins.
\newblock {Bandit processes and dynamic allocation indices}.
\newblock {\em Journal of the Royal Statistical Society, Series B
  (Methodological)}, 41(2):148--177, 1979.

\bibitem{granmo2018tsetlin}
O.-C. Granmo.
\newblock The {T}setlin {M}achine - {A} {G}ame {T}heoretic {B}andit {D}riven
  {A}pproach to {O}ptimal {P}attern {R}ecognition with {P}ropositional {L}ogic.
\newblock {\em arXiv:1804.01508}, Apr 2018.

\bibitem{Granmo2007c}
O.-C. Granmo and N.~Bouhmala.
\newblock {Solving the Satisfiability Problem Using Finite Learning Automata}.
\newblock {\em International Journal of Computer Science and Applications},
  4(3):15--29, 2007.

\bibitem{Granmo2010}
O.~C. Granmo and B.~J. Oommen.
\newblock {Optimal sampling for estimation with constrained resources using a
  learning automaton-based solution for the nonlinear fractional knapsack
  problem}.
\newblock {\em Applied Intelligence}, 33(1):3--20, 2010.

\bibitem{Granmo2010g}
O.-C. Granmo and B.~J. Oommen.
\newblock {Solving Stochastic Nonlinear Resource Allocation Problems Using a
  Hierarchy of Twofold Resource Allocation Automata}.
\newblock {\em IEEE Transactions on Computers}, 59(4):545--560, 2010.

\bibitem{Granmo2007d}
O.-C. Granmo, B.~J. Oommen, S.~A. Myrer, and M.~G. Olsen.
\newblock {Learning Automata-based Solutions to the Nonlinear Fractional
  Knapsack Problem with Applications to Optimal Resource Allocation}.
\newblock {\em IEEE Transactions on Systems, Man, and Cybernetics, Part B},
  37(1):166--175, 2007.

\bibitem{Haugland2014}
V.~Haugland, M.~Kj{\o}lleberg, S.-E. Larsen, and O.-C. Granmo.
\newblock {A two-armed bandit collective for hierarchical examplar based mining
  of frequent itemsets with applications to intrusion detection}.
\newblock {\em Transactions on Computational Collective Intelligence XIV},
  8615:1--19, 2014.

\bibitem{hauser13}
J.~R. Hauser, O.~Toubia, T.~Evgeniou, R.~Befurt, and D.~Dzyabura.
\newblock {D}isjunctions of {C}onjunctions, {C}ognitive {S}implicity, and
  {C}onsideration {S}ets.
\newblock {\em Journal of Marketing Research}, 47(3):485--496, 2010.

\bibitem{hu2018squeeze}
J.~Hu, L.~Shen, and G.~Sun.
\newblock Squeeze-and-excitation networks.
\newblock In {\em Proceedings of the IEEE conference on computer vision and
  pattern recognition}, pages 7132--7141, 2018.

\bibitem{Lammie2019}
C.~{Lammie}, W.~{Xiang}, and M.~{Rahimi Azghadi}.
\newblock {Accelerating Deterministic and Stochastic Binarized Neural Networks
  on FPGAs Using OpenCL}.
\newblock {\em arXiv:1905.06105}, May 2019.

\bibitem{lecun1998gradient}
Y.~LeCun, L.~Bottou, Y.~Bengio, P.~Haffner, et~al.
\newblock Gradient-based learning applied to document recognition.
\newblock {\em Proceedings of the IEEE}, 86(11):2278--2324, 1998.

\bibitem{LiangAAAI19}
Y.~Liang and G.~Van~den Broeck.
\newblock Learning logistic circuits.
\newblock In {\em Proceedings of the 33rd Conference on Artificial Intelligence
  (AAAI)}, jan 2019.

\bibitem{Liu2018}
R.~Liu, J.~Lehman, P.~Molino, F.~P. Such, E.~Frank, A.~Sergeev, and
  J.~Yosinski.
\newblock An intriguing failing of convolutional neural networks and the
  coordconv solution.
\newblock In {\em Proceedings of the 32Nd International Conference on Neural
  Information Processing Systems}, NIPS'18, pages 9628--9639, 2018.

\bibitem{mccormick15}
T.~McCormick, C.~Rudin, and D.~Madigan.
\newblock {A} {H}ierarchical {M}odel for {A}ssociation {R}ule {M}ining of
  {S}equential {E}vents: {A}n {A}pproach to {A}utomated {M}edical {S}ymptom
  {P}rediction.
\newblock {\em Annals of Applied Statistics}, 2011.

\bibitem{Narendra1989}
K.~S. Narendra and M.~A.~L. Thathachar.
\newblock {\em {Learning Automata: An Introduction}}.
\newblock Prentice-Hall, Inc., 1989.

\bibitem{Oommen2007a}
B.~Oommen, S.~Misra, and O.-C. Granmo.
\newblock {Routing bandwidth-guaranteed paths in MPLS traffic engineering: A
  multiple race track learning approach}.
\newblock {\em IEEE Transactions on Computers}, 56(7), 2007.

\bibitem{Oommen1997}
B.~J. Oommen.
\newblock {Stochastic Searching on the Line and its Applications to Parameter
  Learning in Nonlinear Optimization}.
\newblock {\em IEEE Transactions on Systems, Man, and Cybernetics, Part B},
  27(4):733--739, 1997.

\bibitem{Oommen2008}
B.~J. Oommen, S.-W. Kim, M.~T. Samuel, and O.-C. Granmo.
\newblock {A solution to the stochastic point location problem in metalevel
  nonstationary environments}.
\newblock {\em Systems, Man, and Cybernetics, Part B: Cybernetics, IEEE
  Transactions on}, 38(2):466--476, 2008.

\bibitem{Oommen1988}
B.~J. Oommen and D.~C. Ma.
\newblock {Deterministic Learning Automata Solutions to The Equipartitioning
  Problem}.
\newblock {\em IEEE Transactions on Computers}, 37(1):2--13, 1988.

\bibitem{Peharzetal18}
R.~Peharz, A.~Vergari, K.~Stelzner, A.~Molina, M.~Trapp, K.~Kersting, and
  Z.~Ghahramani.
\newblock Probabilistic deep learning using random sum-product networks.
\newblock 2018.

\bibitem{phoulady2019weighted}
A.~{Phoulady}, O.-C. {Granmo}, S.~R. {Gorji}, and H.~A. {Phoulady}.
\newblock {The Weighted Tsetlin Machine: Compressed Representations with Clause
  Weighting}.
\newblock In {\em Proceedings of the Ninth International Workshop on
  Statistical Relational AI (StarAI 2020)}, 2020.

\bibitem{rastegari2016xnor}
M.~Rastegari, V.~Ordonez, J.~Redmon, and A.~Farhadi.
\newblock Xnor-net: Imagenet classification using binary convolutional neural
  networks.
\newblock In {\em European Conference on Computer Vision}, pages 525--542.
  Springer, 2016.

\bibitem{Robbins1952}
H.~Robbins.
\newblock {Some aspects of the sequential design of experiments}.
\newblock {\em Bulletin of the American Mathematical Society}, 1952.

\bibitem{Rudin2019}
C.~Rudin.
\newblock {Stop explaining black box machine learning models for high stakes
  decisions and use interpretable models instead}.
\newblock {\em Nature Machine Intelligence}, 1(5):206--215, 2019.

\bibitem{rudin13}
C.~Rudin, B.~Letham, and D.~Madigan.
\newblock Learning theory analysis for association rules and sequential event
  prediction.
\newblock {\em Journal of Machine Learning Research}, 14:3441--3492, 2013.

\bibitem{Tsetlin1961}
M.~L. Tsetlin.
\newblock {On behaviour of finite automata in random medium}.
\newblock {\em Avtomat. i Telemekh}, 22(10):1345--1354, 1961.

\bibitem{Tung1996}
B.~Tung and L.~Kleinrock.
\newblock {Using Finite State Automata to Produce Self-Optimization and
  Self-Control}.
\newblock {\em IEEE Transactions on Parallel and Distributed Systems},
  7(4):47--61, 1996.

\bibitem{valiant12}
L.~G. Valiant.
\newblock {A} {T}heory of the {L}earnable.
\newblock {\em Communications of the ACM}, 27(11):1134--1142, 1984.

\bibitem{wang6}
T.~Wang, C.~Rudin, F.~Doshi-Velez, Y.~Liu, E.~Klampfl, and P.~MacNeille.
\newblock {A} {B}ayesian {F}ramework for {L}earning {R}ule {S}ets for
  {I}nterpretable {C}lassification.
\newblock {\em The Journal of Machine Learning Research}, 18(1):2357--2393,
  2017.

\bibitem{wheeldon2020hardware}
A.~{Wheeldon}, R.~{Shafik}, A.~{Yakovlev}, J.~{Edwards}, I.~{Haddadi}, and
  O.-C. {Granmo}.
\newblock {Tsetlin Machine: A New Paradigm for Pervasive AI}.
\newblock In {\em Proceedings of the SCONA Workshop at Design, Automation and
  Test in Europe (DATE)}, 2020.

\bibitem{xiao2017}
H.~Xiao, K.~Rasul, and R.~Vollgraf.
\newblock Fashion-mnist: a novel image dataset for benchmarking machine
  learning algorithms.
\newblock {\em arXiv:1708.07747}, 2017.

\bibitem{Yazidi2012b}
A.~Yazidi, O.-C. Granmo, and B.~Oommen.
\newblock {Service selection in stochastic environments: A learning-automaton
  based solution}.
\newblock {\em Applied Intelligence}, 36(3):617--637, 2012.

\bibitem{Yazidi2013}
A.~Yazidi, O.-C. Granmo, and B.~Oommen.
\newblock {Learning-Automaton-Based Online Discovery and Tracking of
  Spatiotemporal Event Patterns}.
\newblock {\em IEEE Transactions on Cybernetics}, 43(3):1118 -- 1130, 2013.

\bibitem{Yazidi2018}
A.~Yazidi and B.~{John Oommen}.
\newblock {On the analysis of a random walk-jump chain with tree-based
  transitions and its applications to faulty dichotomous search}.
\newblock {\em Sequential Analysis}, 37:31--46, Jan 2018.

\bibitem{zeiler2014visualizing}
M.~D. Zeiler and R.~Fergus.
\newblock Visualizing and understanding convolutional networks.
\newblock In {\em European conference on computer vision}, pages 818--833.
  Springer, 2014.

\bibitem{Zhang2016a}
J.~Zhang, Y.~Wang, C.~Wang, and M.~Zhou.
\newblock {Symmetrical Hierarchical Stochastic Searching on the Line in
  Informative and Deceptive Environments}.
\newblock {\em IEEE Transactions on Cybernetics}, 47(3):626 -- 635, Jul 2016.

\end{thebibliography}

\end{document}